\documentclass[conference]{IEEEtran}
\IEEEoverridecommandlockouts
\usepackage{cite}
\usepackage{amsmath,amssymb,amsfonts}
\usepackage{algorithmic}
\usepackage{graphicx}
\usepackage{textcomp}
\usepackage{xcolor}
\usepackage{soul}
\usepackage{csquotes}
\def\BibTeX{{\rm B\kern-.05em{\sc i\kern-.025em b}\kern-.08em
    T\kern-.1667em\lower.7ex\hbox{E}\kern-.125emX}}

\usepackage{makecell}

\newcommand\topic[1]{\textcolor{blue}{}}

\begin{document}

\title{Understanding Concept Identification as Consistent Data Clustering Across Multiple Feature Spaces} 

\author{\IEEEauthorblockN{Felix Lanfermann}
\IEEEauthorblockA{\textit{Honda Research Institute Europe} \\
Offenbach, Germany \\
felix.lanfermann@honda-ri.de}
\and
\IEEEauthorblockN{Sebastian Schmitt}
\IEEEauthorblockA{\textit{Honda Research Institute Europe} \\
Offenbach, Germany \\
sebastian.schmitt@honda-ri.de}
\and
\IEEEauthorblockN{Patricia Wollstadt}
\IEEEauthorblockA{\textit{Honda Research Institute Europe} \\
Offenbach, Germany \\
patricia.wollstadt@honda-ri.de}
}

\maketitle

\begin{abstract}
Identifying meaningful concepts in large data sets can provide valuable insights into engineering design problems.
Concept identification aims at identifying non-overlapping groups of design instances that are similar in a joint space of all features, but which are also similar when considering only subsets of features.
These subsets usually comprise features that characterize a design with respect to one specific context, for example, constructive design parameters, performance values, or operation modes.
It is desirable to evaluate the quality of design concepts by considering several of these feature subsets in isolation. 
In particular, meaningful concepts should not only identify dense, well separated groups of data instances, but also provide non-overlapping groups of data that persist when considering pre-defined feature subsets separately.
In this work, we propose to view concept identification as a special form of clustering algorithm with a broad range of potential applications beyond engineering design.
To illustrate the differences between concept identification and classical clustering algorithms, we apply a recently proposed concept identification algorithm to two synthetic data sets and show the differences in identified solutions.
In addition, we introduce the mutual information measure as a metric to evaluate whether solutions return consistent clusters across relevant subsets.
To support the novel understanding of concept identification, we consider a simulated data set from a decision-making problem in the energy management domain and show that the identified clusters are more interpretable with respect to relevant feature subsets than clusters found by common clustering algorithms and are thus more suitable to support a decision maker.

\end{abstract}


\section{Introduction}

\topic{Concept Identification solves specific problems in the engineering design domain}
In the engineering design domain, the identification of \emph{design concepts} is a central
task that identifies groups of engineering designs that share similar
characteristics, typically in terms of their specification, but also in terms of performance 
or other types of descriptions, such as operation mode. 
When identifying concepts, it is relevant that identified groups of designs are similar with
respect to \emph{all} describing features, but 
that this similarity is also preserved when considering a subset of feature types in isolation.
For example, designs within a concept should be highly similar with respect to all characteristics, 
but designs should also be similar in terms of only their specifications or only their performance. 
This not only creates insight into the relation between the distribution of design groups and their
corresponding quality criteria.
It is further relevant for identifying highly representative instances of design concepts, also termed 
\emph{archetypes}, which may then be used as a starting point for further refinement of
specifications, which lead to consistent performance.

\topic{We approach Concept identification as a clustering problem with specific constraints on the generated clusters}
\topic{The solution to some problems are non-overlapping and consistent clusters in arbitrarily choosable description spaces}
\topic{Description spaces are ...}
Concept identification may be viewed as a special form of clustering problem that aims at identifying 
non-overlapping groups of instances that have high similarity in a joint feature space, but which persist
when considering relevant subspaces of features. A subspace is here defined as an arbitrary subset of features that defines an instance in 
a specific context or domain, into which instances are projected from the joint feature space.
Hence, concept identification may be approached using a clustering algorithm with specific
constraints on the returned solution. In particular, the solution should a) comprise dense clusters 
in the joint space of all features, b) the clusters should be non-overlapping, which can only be achieved for some data sets by not assigning 
all instances to a cluster, and c), non-overlapping clusters
should persist in relevant subspaces of the joint space. In other words, a suitable
algorithm should return non-overlapping clusters in the joint feature space and instances that 
are assigned to a cluster in the joint space should be assigned to the same cluster in
arbitrary, a-priori defined subspaces. We term this later property \emph{consistency} of clusters across
different feature spaces. As a result of this consistency constraint, instances being grouped within 
the same cluster, show high similarity both in the joint and in the individual subspaces. This further leads to also 
a high similarity \emph{between} subspaces for instances assigned to concepts.

\topic{Our algorithm developed for Concept identification can be applied as a clustering technique for other clustering problems}
\topic{we have published a concept identification algorithm. In this paper we use MI to
show: a) Concept identification leads to consistent clusters in subspaces and b)
existing clustering algorithm do not have this property.}
\topic{We choose MI over traditional cluster quality metrics, because traditional metrics do not capture, or even conflict with the targets of the approach.}
We here evaluate a recently proposed algorithm based on evolutionary optimization that 
was introduced to perform concept identification in the engineering design domain
\cite{Lanfermann2020,Lanfermann2022}, in terms of a general clustering algorithm which we 
believe to be relevant also to other application domains. 
In particular, we show that the concept identification algorithm returns clustering solutions with the 
properties introduced above and compare them to solutions returned by classical clustering 
algorithms. We highlight the differences between the two approaches, in particular, the ability 
of the concept identification algorithm to identify consistent clusters across arbitrary, predefined subspaces, a property that may 
be relevant also in other application domains, for example, cross-domain recommender systems
\cite{Khan2017}. We here demonstrate such an application to a decision making problem from 
the energy management domain.

We evaluate the concept identification algorithm's ability to return solutions according to the three constraints defined above.
To evaluate the algorithm's ability to identify clusters in subspaces with high 
similarity between clusters, as enforced by the consistency constraint, we use the mutual information (MI) to evaluate returned solutions, 
since traditional metrics for evaluating clustering solutions do not account for this
property. 

\topic{This paper is structured as follows..}
\topic{We demonstrate the approach with experiments}
The remainder of this paper is structured as follows: In Section~\ref{sec:methods}, we review prior art
in clustering and concept identification. We then briefly introduce the algorithm evaluated here and
highlight how it differs from existing clustering approaches. Last, we introduce the MI
for evaluation of clustering solutions. In Section~\ref{sec:experiments}, we
apply the algorithm to three artificial data sets to demonstrate the properties of
clustering solutions returned by the algorithm. The first two experimental data sets are
designed to highlight the algorithm's ability to identify dense, non-overlapping clusters that are consistent
across subspaces. The third data set is generated through high-fidelity simulation and illustrates a
use case for the algorithm in a decision-making problem from the energy management domain. In Section~IV, we discuss
our findings and potential relevant application scenarios of the algorithm. We close in
Section~V with a conclusion and an outlook on future work.

\section{Methods}\label{sec:methods}

\subsection{Concept identification as an extended form of clustering}

\topic{CI is a problem known in the engineering domain and concern the grouping of design solutions and their performance criteria under a set of constraints.}

\topic{The constraints are compactness, separation (non-overlapping), and consistency (high correlation between projections into arbitrary subspaces.)} 
In short, there are three objectives that should be met by a suitable solution for concept identification, namely a) compactness of clusters, b) no overlap between clusters, and c) a consistency of clusters that persists in arbitrary subspaces of the joint clustering space.
The first two objectives are (partially) optimized by classical clustering algorithms \cite{Hastie2009}, i.e., algorithms that find a \enquote{grouping} or assignment of samples to \enquote{clusters} that results in a low variance or \emph{compactness} within clusters and a high \emph{separation} between clusters \cite{Liu2010}. 
Hence, existing clustering algorithms have objectives that are related to those of concept identification, but differ in one aspect: 
Clustering does not ensure the third objective of consistent clusters in arbitrary subspaces. 
There are extensions of the classical clustering objective, which add additional constraints on clustering solutions, e.g., subspace \cite{Parsons2004} or multi-view clustering \cite{Bickel2004}. 
However, none of these approaches return solutions that are suitable to solve the problem of concept identification. 
We will discuss these existing approaches and their difference to concept identification below.

Before we discuss the relationship between concept identification and clustering in the next subsection, we will briefly review existing concept identification approaches and describe the algorithm, originally introduced in \cite{Lanfermann2020}.

Concept identification algorithms typically define a \emph{quality metric} to asses whether cluster assignments in the chosen 
subspaces follow the constraints defined above.
The quality can, for example, be determined by calculating the interestingness and significance (IS)~\cite{Tan2000} of assignments.
This produces a numerical, therefore comparable metric to compare different data sets.
Another approach to assess the quality of concepts via a quality metric is used in multi-objective optimization~\cite{Graening2014}.
Here, concepts are judged by how well they cover the pareto-front of optimal solutions.
In particular, the hyper-volume of the concepts is used to judge their quality.
However, this approach is only valid to assess concepts in data generated by multi-objective optimizations and is not generally applicable to data sets from other domains.

Although both measures provide a useful estimate of concept quality, they cannot be used within a concept identification algorithm that is applied to return solutions according to the constraints described above, as these measures lack the ability to evaluate multiple concepts simultaneously. 
The measures cannot explicitly consider and penalize overlap of concepts within the subspaces and do not evaluate a concept's extent.
Identifying concepts by optimizing the referenced metrics can therefore lead to trivial solutions given as either indistinguishable concepts that span the entire data set (as this would be an optimal assignment), or as insignificantly small concepts containing a single point of data only.

The identification of concepts is nevertheless possible by penalizing concept overlap in an arbitrary number of subspaces, as well as very large and very small concepts.
A concept quality metric that integrates both of these constraints is proposed by \cite{Lanfermann2022}.
Here, a data set
\begin{equation}\label{eq:dataset}
    \mathcal{D} = \left\{x_i, i = 1, ...,  N_\mathcal{D} \right\}
\end{equation}
with $N_{D}$ samples is considered. 
Each sample
\begin{align}
    x=(x^1,\dots,x^{N_{S}}, x^\Delta)^T\,
\end{align}
is represented by features that are divided into $N_{S}$ subspaces.
Any features that are not associated with a subspace are denoted by $x^\Delta$.

For every concept $\alpha=1,\dots,N_{C}$, the approach defines a connected region $C_{\alpha, k}$ in each subspace $k=1,\dots,N_{S}$.
All samples that lie within this region are considered candidates for the respective concept.
A concept is then defined as the term
\begin{equation}\label{eq:concept_c_alpha}
    C_\alpha = \Big\{  x \in \bigcap_{k=1}^{N_{S}} C_{\alpha, k} \Big| x \not \in C_{\beta, k} ; \forall k,\beta\neq\alpha\Big\},
\end{equation}
where $k,l=1,\dots,N_{S}$ enumerate the feature subspaces and  $\alpha,\beta=1,\dots, N_C$ enumerate the concepts. 
The concept $C_{\alpha}$ therefore only contains those samples, that lie within all regions $\{C_{\alpha, 1}, \dots, C_{\alpha, N_{S}}\}$, but not within any of the regions $\{C_{\beta, 1}, \dots, C_{\beta, N_{S}}\}$ associated with any other concept $\beta$.

The approach proposes a \emph{concept quality metric} (CQM)
\begin{equation}\label{eq:q}
 Q=\prod_\alpha^{N_C}Q_\alpha\:,
\end{equation}
where the individual metric $Q_\alpha$ for each concept is defined as 
\begin{equation}\label{eq:q_alpha}
 \begin{aligned}
  Q_\alpha = & \left(\prod_{k}^{N_{S}} \sqrt[N_{S}]{\frac{| C_\alpha |} {|C_{\alpha,k} |}} \right)\cdot F \left( \frac{|C_{\alpha} |} {N_\mathcal{D}}, s \right) \\
  & \cdot F \left( \frac{|P_{\alpha}|} {|P| }, p \right)\:.
  \end{aligned}
\end{equation}

For the second and third term in \eqref{eq:q_alpha}, a scaling function
\begin{equation}
\label{eq:scaling_function}
  F(x,y) = \begin{cases}
			\sqrt{1 - \left( \frac{x - y}{y} \right) ^2}, & \text{if $x < y$,}\\
			\quad 1, & \text{if $ y < x < 1-y$}, \\
			\sqrt{1 - \left( \frac{x -1 + y}{y} \right) ^2}, & \text{if $x > 1-y$}
		\end{cases}
\end{equation}
is utilized.
The second term penalizes concepts that contain fewer or more samples of the data set than a predefined value $s$.
A optional set of samples $P$ can be defined to further steer the identification process towards user-preference.
As this option was not used in the conducted experiments in section~\ref{sec:experiments}, we refer to \cite{Lanfermann2022} for more details.

The CQM in \eqref{eq:q} quantifies the quality of a given distribution of concepts.
It can therefore be used to identify an optimal set of concepts by maximizing the CQM in a numerical optimization problem.

One option is to apply the metric as the evaluation function in an evolutionary optimization framework as suggested by \cite{Lanfermann2020}, where regions of potential concept candidates are adapted to find an optimal distribution.
In an $n$-dimensional subspace of the dataset, such a region can be represented by one $n$-dimensional hyper-ellipsoid.
The hyper-ellipsoid has 
\begin{equation}\label{eq:num_par}
N_{P,S} = n(n+3)/2
\end{equation}
parameters (for each $n$-dimensional subspace) which need to be optimized by the evolutionary search algorithm.

The approach from \cite{Lanfermann2022} is implemented and used within the experimental evaluations in section \ref{sec:experiments}.

\subsection{Difference to existing clustering approaches}

\topic{Existing clustering methods do not generate clusters with the properties required in CI} 
As introduced above, clustering algorithms in their most general form target a grouping of data that yields compact and well-separated clusters \cite{Hastie2009}. 

Well-known examples of such approaches are k-means \cite{MacQueen1967} or k-means++ \cite{Arthur2007}.
Both approaches divide all data samples into a predefined number of clusters. However, solutions are not necessarily non-overlapping, and 
clusters found in the joint feature space do not lead to well-defined clusters in the defined subspaces. Also, these clustering 
approaches assign all instances to a cluster, which is not a requirement for concept identification and may even prevent the 
finding of a suitable solution \cite{Lanfermann2022}.

Fuzzy c-means~\cite{Bezdek1981} and Gaussian mixture models~\cite{Rasmussen2000} allow for a less strict allocation of samples to clusters and thus 
relax the strict assignment of instances. However, these algorithms do also not lead to consistent clusters across subspaces.
The approaches cannot identify corresponding groups in projections of the full feature vector simultaneously, in which each cluster in one projection corresponds to one cluster in all other projections.
These clustering approaches are thus not suited to solve the defined problem, as they cannot be easily extended or adapted to realize the required constraints on the solutions \cite{Lanfermann2020}. 

In contrast to the traditional clustering approaches discussed above, subspace clustering~\cite{Parsons2004} identifies clusters in lower-dimensional projections when applied on high-dimensional data.
This is particularly beneficial when aiming at data-driven dimensionality reduction and mitigating sparsity of the data.
Nonetheless, the user cannot predetermine the subspaces, into which instances are projected in advance.
However, such an alignment of the projections within user-defined subspaces would be necessary to solve concept identification problems, where 
subspaces represent specific domains or types of features and thus have semantic meaning that should be preserved.
For this reason, and since subspace clustering might (intentionally) lead to overlapping clusters~\cite{Sim2013}, the approach is not applicable for the defined problem.

Information maximization clustering (IMC) \enquote{simultaneously learns discriminative representations while generating labels in an unsupervised mode}~\cite{Ntelemis2022}.
Similar to subspace clustering, the representations are learned within the process and no a-priori defined subspaces are taken into consideration.
The desired consistency of groups across subspaces can hence also not be assured.

The listed approaches may return compact clusters that are separable with respect to the considered features.
However, consistency of the clusters with respect to multiple subspaces of the features is not achieved.
There are, however, other approaches that integrate a notion of consistency into the clustering process.
Often, a data set is constructed from different \enquote{views} on the data, where a view can be understood as a group of connected features. 
The heterogeneous properties of the data might hold a potential connection~\cite{Yang2018} and the views provide complementary information about the data which can be exploited~\cite{Yang2018}.
Here, a view may be understood as a subspace, such that---in accordance with our constraints defined above---multi-view clustering~\cite{Bickel2004} maximizes a cluster quality metric within each view, while accounting for clustering consistency across different views~\cite{Yang2018} (subspaces, respectively).
High consistency can, for example, be achieved by alternating the maximization and expectation step for each view of the data~\cite{Bickel2004}, or optimizing a weight factor for each view in a k-means-based clustering loss-function~\cite{Cai2013}.
Combining multi-view and subspace clustering can uncover clusters in concealed subspaces in multi-view data \cite{Gao2015}.
However, multiple views are utilized in the process to optimize clustering performance, not to achieve highest possible consistencies across all views.
Consistency is factored in as a trade-off relation, but overlapping clusters in individual views are not fully avoidable.
This is, nevertheless, a central requirement in concept identification and renders the usage of multi-view clustering for concept identification difficult.

\subsection{Evaluation of consistency with mutual information}

\topic{We use Mutual information to validate our algorithm and compare it to other algorithms, because traditional cluster quality metrics are conflicting with constraints.} As introduced in the last subsection, the concept identification algorithm evaluated here differs from existing clustering algorithms by returning strictly non-overlapping clusters and by ensuring consistency of clusters in subspaces. To evaluate whether these objectives are met, we compare the algorithm to existing clustering algorithms and evaluate solutions by, first, calculating the Silhouette Coefficient~\cite{Rousseeuw1987} to demonstrate that the concept identification algorithm returns compact clusters, with a similar separation than found in clusters returned by existing clustering algorithms. Second, we apply the mutual information (MI), a measure from information theory \cite{Shannon1948}, to evaluate the consistency criterion. Note that existing metrics for validating clustering solutions quantify compactness and separation, the objectives of classical clustering algorithms \cite{Liu2010}. Examples of these metrics are the Silhouette Coefficient as used here or the Davies-Bouldin index \cite{Davies1979}. However, these metrics are not suitable to judge the third criterion of concept identification, the consistency of clusters across subspaces.

\topic{Why do we expect high similarity between subspaces?} 
\topic{MI shows consistency between subspaces} The central difference between concept identification and regular clustering algorithms is 
the enforced consistency of clusters between the joint and all a-priori defined subspaces. As a result of this consistency, instances that are 
assigned to a cluster show a high similarity in the joint space, but also in all subspaces. This similarity within subspaces should lead to 
a high similarity also \emph{between} subspaces. In other words, when considering only one subspace, knowing that an instance is assigned to 
a specific cluster, is informative about the assignment of that instance in all other subspaces. Accordingly, the properties of an instance in 
one subspace are informative about the properties of that instance in another subspace.

Hence, we here evaluate the similarity between subspaces that is enforced by the consistency of clusters, using the MI. MI quantifies how much information one variable provides about a second. Thus, it can be used to measure how informative an instance's value in one subspace is about the same instance’s value in another subspace. Hence, we calculate the MI between targeted subspaces for all instances assigned to a cluster, to quantify if the algorithm satisfies our requirement for producing clusters where instances behave similarly in subspaces of interest.

\topic{This is how MI works} The MI quantifies the amount of information one random variable, $X$, provides about a second random variable, $Y$, \cite{MacKay2005} as

\begin{equation}
  \begin{aligned}
        I(X;Y) =&  \sum_{x,y} p(x,y) \log \frac{p(x|y)}{p(x)} = \sum_{x,y} p(x,y) \log \frac{p(x,y)}{p(x)p(y)},
    \label{eq:mi}
  \end{aligned}
\end{equation}

\noindent where we write $p(x)$ as a shorthand for the probability $p(X=x)$ of variable $X$ taking on the value $x$.
The MI may be interpreted as the Kullback–Leibler divergence between the product of the marginal distributions, $p(x)p(y)$, and the joint distribution, $p(x,y)$, and thus measures the deviation of the two variables' joint distribution from statistical independence.

We estimate the MI using a nearest-neighbor-based estimator for continuous data proposed by \cite{Kraskov2004} and implemented in the IDTxl Python toolbox \cite{Wollstadt2019}, which uses an estimator from the JIDT toolbox \cite{Lizier2014}. 
Note that it is a well-known problem that MI estimates from finite data suffer from estimation bias (e.g., \cite{Paninski2003}). 
One approach to handle estimation bias in continuous estimators is the use of permutation testing. 
Here, we generate a Null-distribution representing the Null-hypothesis of $X$ and $Y$ being statistically independent by repeatedly shuffling one of the two variables and estimating the MI from this shuffled data. 
The original estimate is then compared against this distribution and a $p$-value is calculated as the fraction of estimates from shuffled data that are higher than the original estimate (see, e.g., \cite{Vicente2011}).

\section{Experiments}\label{sec:experiments}
\topic{we did three experiment: trivial, not trivial, application}
A set of experiments was conducted to illustrate and quantify the differences between concept identification and clustering approaches.
In each experiment, various clustering techniques, as well as the introduced concept identification algorithm were applied to a data set.
The first two experiments were conducted on artificial data sets to illustrate the specific properties of solutions returned by the applied concept identification algorithm.
The third experiment compares the results of one clustering technique and the referenced concept identification approach on an application data set from the field of energy management in a decision-making problem. 
Table~\ref{tab:experiment_setup} provides an overview of the utilized data sets and clustering preliminaries.

\begin{table}[ht]
\centering
\caption{Details of the experimental data}
\begin{tabular}{c c c c}
\hline
Experiment & 2D & 4D & 3D \\
\hline
data source & artificial & artificial & \makecell{building simulations} \\
\# data samples & 34000 & 30000 & 20699 \\
\# subspaces $N_{S}$ & 2 & 2 & 2\\
\# total features & 2 & 4 & 3\\
\# features per space & 1, 1 & 2, 2 & 1, 2 \\
\# clusters / concepts $N_{C}$  & 3 & 3 & 3 \\
\# opt. parameters $N_{P}$ & 12 & 30 & 21\\
\hline
\end{tabular}
\label{tab:experiment_setup}
\end{table}

\subsection{Two-dimensional artificial data}

\begin{figure*}[ht]
    \centering
    \includegraphics[width=0.96\textwidth]{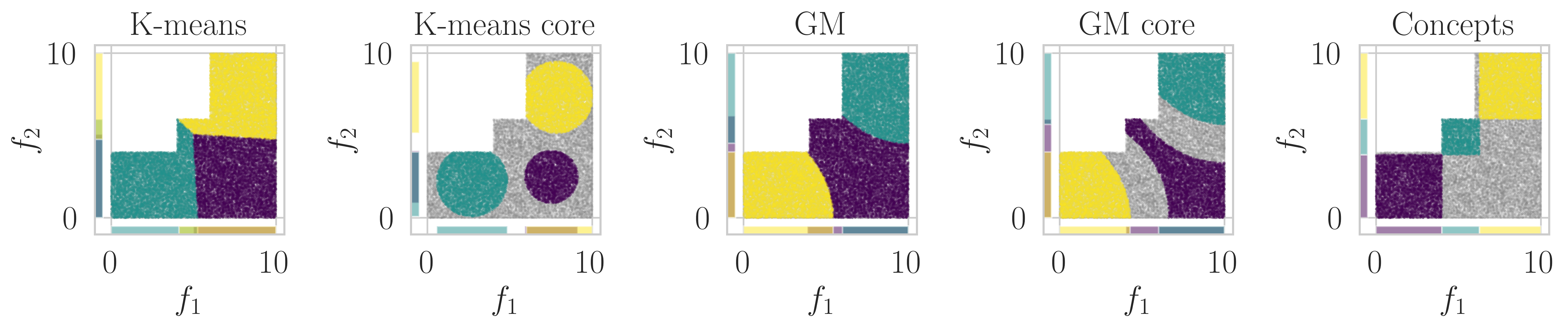}
    \caption{Partitions of the data set into clusters and concepts based on different methods: Four clustering and one concept identification method are applied to an artificial two-dimensional data set to divide the data into consistent and compact groups. The goal to derive a separation into groups that do not overlap when the data set is projected onto one a single feature is achieved only by the concept identification approach. The clusters and concepts are represented by the purple, green and yellow samples. The projections of the clusters and concepts onto the features are depicted next to the axes.}
    \label{fig:MI_2dim}
\end{figure*}

\topic{To illustrate the clustering solutions and their properties, we create a simple synthetic data set and applied multiple clustering algorithms and concept identification on it.}
To illustrate the main differences between the solutions given by traditional clustering techniques and the evaluated concept identification algorithm, a simple synthetic data set, composed of 34000 samples, was created.
Each point of data is represented by two features, $f_{1}$ and $f_{2}$, (Fig.~\ref{fig:MI_2dim}).
\topic{data generation}
The data set was uniformly sampled from an area given by

\begin{equation}
\label{eq:dataset_2dim}
  A = \begin{cases}
			0 \leq f_1 < 4, & \text{if } 0 \leq f_2 < 4,\\
			4 \leq f_1 < 10, & \text{if } 0 \leq f_2 < 6,\\
			6 \leq f_1 < 10, & \text{if } 6 \leq f_2 < 10.\\
		\end{cases}
\end{equation}

To demonstrate the robustness of the approaches, the experiment was repeated 20 times.

\topic{The task of the experiment}
According to the three constraints defined for the concept identification problem, a 
solution should comprise clusters in the joint space, $(f_1,f_2)$, which are also clearly separable based on subspaces $f_1$ and $f_2$, individually.
It is therefore necessary to find three consistent groups of samples with respect to both feature $f_{1}$ and $f_{2}$, individually, and to the joint feature space. The groups should not overlap when considering either sub- or the joint space.

\topic{k-means and GMM are compared to concept identification}
Four clustering methods were applied to the two-dimensional data set: k-means, Gaussian mixture models (GMM), as well as a modification of each algorithm.
\topic{The modification of the clustering methods is added to make the comparison more fair.}
Since the concept identification approach does not require all samples to be associated with a concept, this might create an unfair advantage for this method compared to clustering algorithms that always assign all instances to a cluster.
Hence, in the modified clustering methods, only those samples were considered to be part of the cluster, that were closest to the respective cluster center\footnote{From the regular k-means method, the maximum inner-cluster distance was calculated. For the modified k-means approach, only those samples, that lay within a radius of 20\% of this maximum distance to the center, were assigned to the cluster. For the modified GMM approach, only those samples, that are allocated to the original groups with a probability rate of larger than 90\%, are assigned to the cluster}.

\topic{The concept identification process uses 2 subspaces with one feature each}
Next, we applied the concept identification algorithm to find groups of samples in the joint spaces, that are consistent within subspaces $f_{1}$ and $f_{2}$.

Since we aimed at identifying three groups in two one-dimensional subspaces, the total number of parameters that needed to be modified by the optimization amounts to $N_{P} = 3 \cdot (1(1+3)/2 + 1(1+3)/2) = 12$, according to \eqref{eq:num_par}.
As the optimization algorithm, a covariance matrix adaptation evolutionary strategy (CMA-ES)\cite{hansenCMA,Hansen2001} was used with a population size of 10 for 1000 generation.

\topic{Interpretation of results}
The different methods lead to very different partitions of the data set (Fig.~\ref{fig:MI_2dim}).
All methods found valid clusters, but only the concept identification method identified groups that meet the required conditions.
If projected onto one of the features, all traditional clustering solutions demonstrated significant overlap between the clusters, showing that none of the clustering algorithms returned solutions that led to consistent clusters in considered subspaces.

\begin{figure}[ht]
    \includegraphics[width=0.45\textwidth]{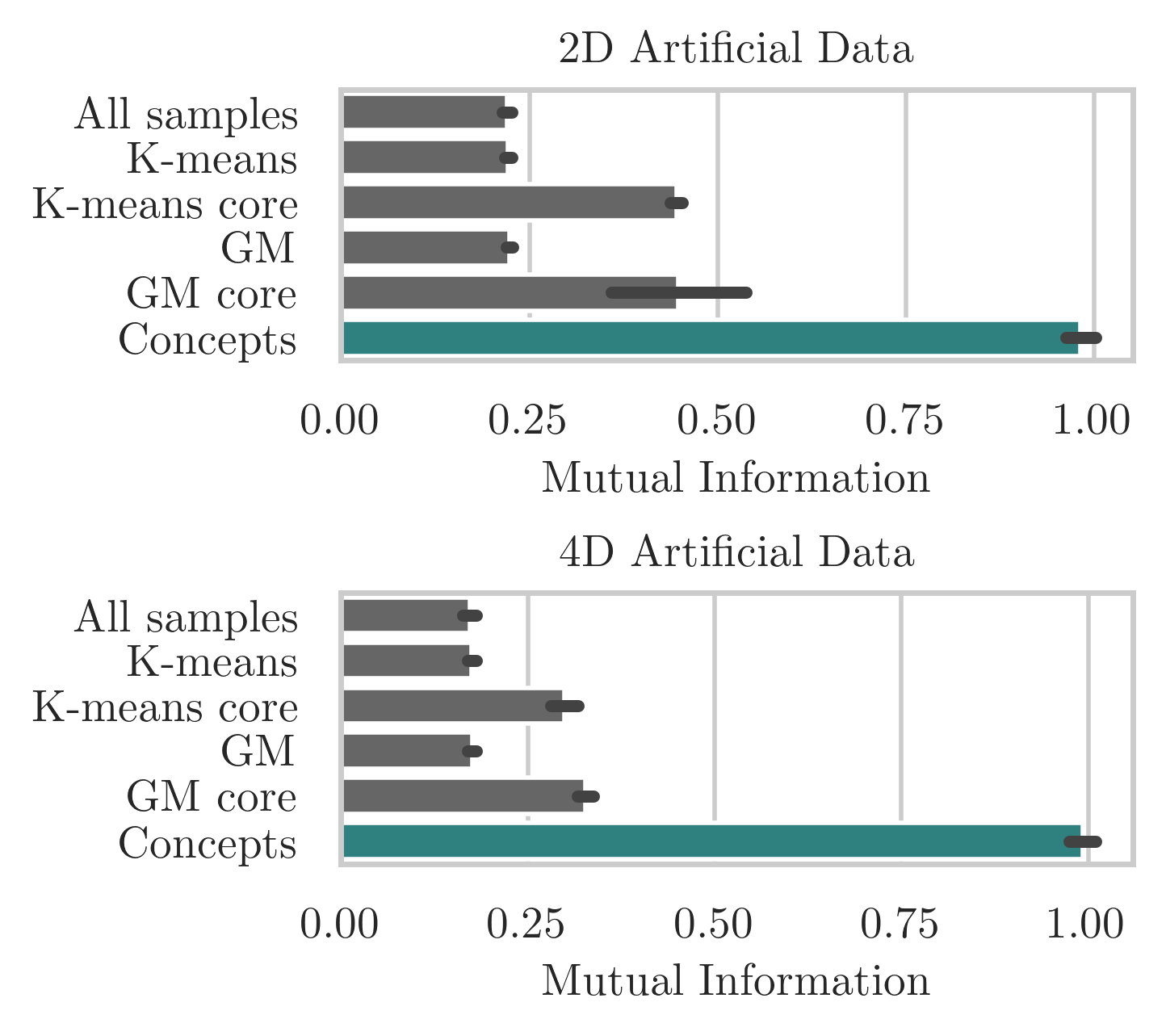}
    \caption{Mutual information (mean and standard deviation for all repeated experiments) between the identified concepts and clusters: The concept identification approach leads to significantly larger MI than the other approaches. The top shows the results for the two-dimensional data set, the bottom shows the results for the four-dimensional data set.}
    \label{fig:MI_bars}
\end{figure}

\topic{The results are quantified with MI}
The results were further evaluated by calculating the MI between both features, $f_1$ and $f_2$, while considering only those samples that were assigned to a cluster. 
\topic{Concept Identification is the best}
Between the clustering approaches, the modified versions of k-means and GMMs led to higher levels of MI, compared to all other traditional clustering algorithms (Fig.~\ref{fig:MI_bars}).
However, the solution found by the concept identification algorithm showed a significantly higher MI than all other solutions. 
Since MI provides a quantitative measure of consistency across subspaces, we conclude that the concept identification algorithm is suited best for assigning instances to clusters such that instances behave highly similar across subspaces.

\begin{figure}[ht]
    \includegraphics[width=0.45\textwidth]{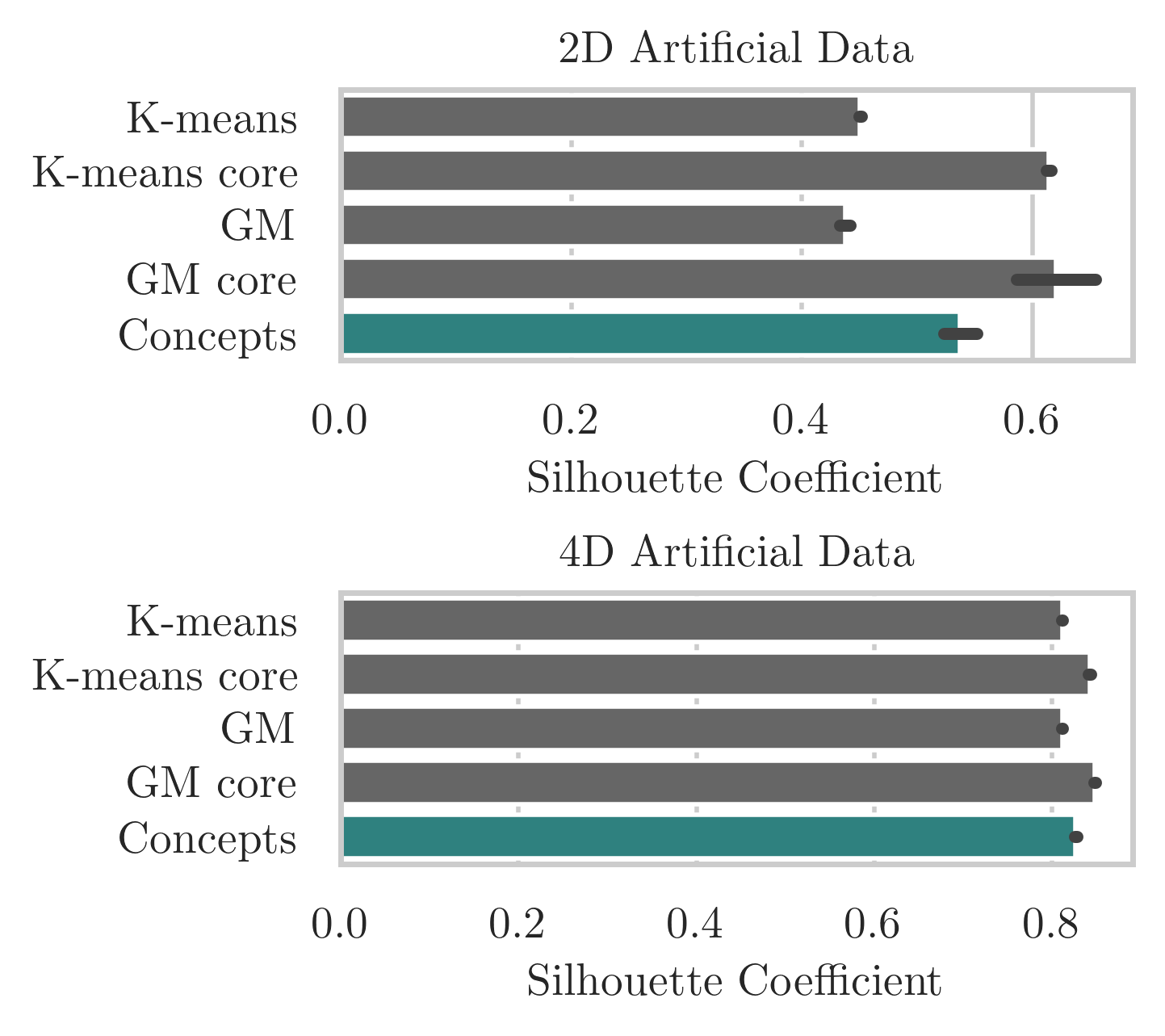}
    \caption{Silhouette Coefficient (mean and standard deviation for all repeated experiments) for the identified concepts and clusters: The concept identification approach and the clustering methods lead to comparable silhouette coefficients.
    The top shows the results for the two-dimensional data set, the bottom shows the results for the four-dimensional data set.}
    \label{fig:SIL_bars}
\end{figure}

To investigate whether the identified concepts also satisfy the need for compactness and separation imposed on common clustering methods, we calculated the silhouette coefficients for all results (Fig.~\ref{fig:SIL_bars}).
Since the clustering method and the concept identification approach lead to comparable scores, we follow that the requirement to produce compact and separable groups is met.
We further emphasize that the concept identification approach leads to non-overlapping groups by definition \eqref{eq:concept_c_alpha}.

\subsection{Four-dimensional artificial data}

\topic{The results of the first experiment look trivial at first glance}
The identified concepts resulting from the first experiment might look trivial at first glance.
In a one-dimensional subspace, the elliptic representation describes each concept by a mean value and a radius.
The distribution of concepts is hence given as a direct segmentation of the one-dimensional data.
Illustrating the clusters in the (full) two-dimensional data set then naturally creates a simple grid structure (Fig.~\ref{fig:MI_2dim}).

\topic{To illustrate that finding clusters with the desired properties becomes non-trivial already in slightly higher dimension spaces, we conduct the second experiment}
In data sets where the subspaces of interest contain more than one dimension, the results are less trivial. 
To demonstrate this effect, we conducted a second experiment involving a second artificial data set.

\topic{The Four-dimensional artificial data is created}
The data set contained 30000 samples, each described by four features $f_1$, $f_2$, $f_3$, and $f_4$, and 
was created by sampling from the normal distribution $\mathcal{N}(\mu_i, \, \sigma^2)$ around three centers $\mu_1 = (0, 0, 0, 0)^T$, $\mu_2 = (10, 10, 10, 10)^T$, and $\mu_3 = (10, 10, 0, 0)^T$.

\topic{The task of this experiment is to identify clusters and concepts that are consistent within two two-dimensional subspaces}
Again, the same four clustering approaches and the concept identification method were applied to identify three groups of samples.
The modified k-means and GMM again only consider those samples to be part of the cluster, that are closest\footnote{From the regular k-means and GMM approaches, the maximum inner-cluster distance is calculated. For the modified approaches, only those samples, that lie within a radius of 10\% of this maximum distance to the center, are assigned to the cluster} to the cluster center.
On the concept identification algorithm, we again imposed the constraint that groups should be consistent within two subspaces.
The first subspace was defined as a combination of features $f_1$ and $f_2$, while the second subspace comprised $f_3$ and $f_4$.
An ideal partition would lead to groups that do not overlap when projected onto the subspaces.

\topic{The concept identification method is done as usual}
For the concept identification approach, each concept is again represented by the combination of one ellipse for each of the two subspaces.
According to \eqref{eq:num_par}, the total number of parameters to be optimized amounts to $N_{P} = 3 \cdot (2(2+3)/2 + 2(2+3)/2) = 30$, since both subspaces are two-dimensional.
The algorithm---again a CMA-ES with a population size of 10 and 1000 generations---hence optimizes the concept distribution by adapting in total 30 parameters.

\begin{figure*}[ht]
    \centering
    \includegraphics[width=0.96\textwidth]{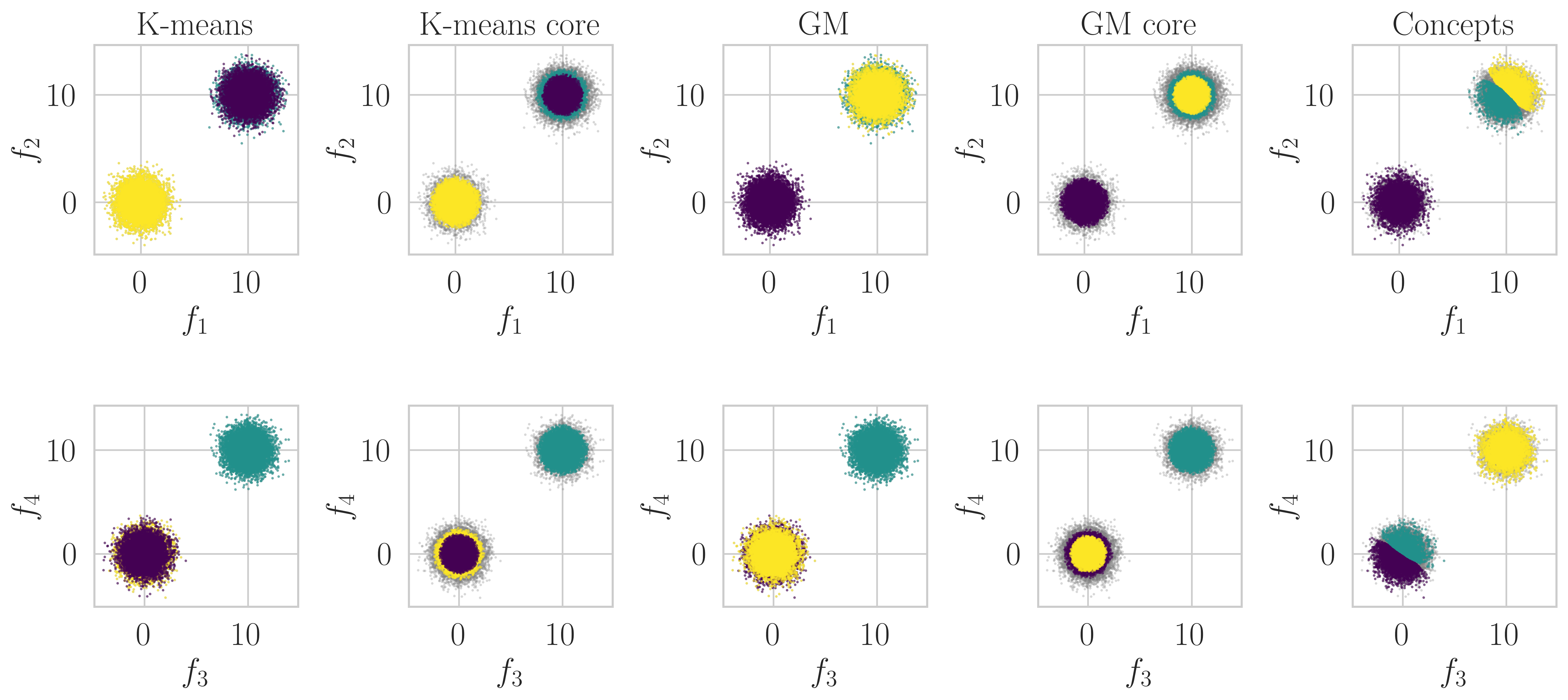}
    \caption{Partitions of the data set into clusters and concepts based on different methods: Four clustering and one concept identification method are applied to an artificial four-dimensional data set to divide the data into consistent and compact groups. The goal to derive a separation into groups that do not overlap when the data set is projected onto the two separate subspaces is achieved only by the concept identification approach. The first subspace is given as the combination of $f_1$ and $f_2$, the second subspace is given as the combination of $f_3$ and $f_4$. The clusters and concepts are represented by the purple, green and yellow samples.}
    \label{fig:MI_4dim}
\end{figure*}

\topic{The results are again quantified with MI and visually evaluated}
The solutions given by all clustering methods show substantial overlap in the individual subspaces (Fig. \ref{fig:MI_4dim}).
All clustering approaches identify sample groups that are distinguishable within the full four-dimensional data set.
The cluster centers are close to the means $\mu_1 = (0, 0, 0, 0)^T$, $\mu_2 = (10, 10, 10, 10)^T$, and $\mu_3 = (10, 10, 0, 0)^T$ that were used to create the data set.
However, when projected into the subspaces, two of the three clusters are indistinguishable. 
In contrast, the groups discovered by the concept identification approach (i.e. the concepts) do not overlap within the subspaces.

\topic{For the problem we have, concept identification is again the best performing approach}
Besides the visual assessment, the results were again evaluated by calculating the MI.
In detail, it is calculated how much information is shared between the concepts and clusters in both subspaces.
Variables $X$ and $Y$ are given as the joint features $f_1$ and $f_2$, as well as $f_3$ and $f_4$, respectively.
As in the previous experiment, the concept identification method found groups with the largest MI (Fig. \ref{fig:MI_bars}), thereby showing that the chosen approach lead to consistent concepts within the two subspaces.

\subsection{Three-dimensional energy management configuration data}

\topic{To show the relevance for real application scenarios, we did an experiment with a realistic use-case}
Last, algorithms were compared on a data set from a realistic application scenario.
\topic{The data set originates from an energy management building configuration optimization}
The data set was created by simulating several thousand building energy management configurations \cite{Liu2022}. 
A common challenge when 
optimizing such configurations is to support a decision maker in selecting an optimal 
solution given the user's specific preferences or requirements. Here, concept 
identification may be used to identify different types or categories of solutions that 
make the selection of a preferable solution easier.

The data set was generated using a many-objective evolutionary algorithm that adapted nine different parameters to achieve optimal configurations, where the quality of an individual configuration was judged by ten partially conflicting objectives.
One configuration was defined by parameters for the size, orientation, and inclination of a photovoltaic (PV) system, the size and operating conditions of a battery energy storage system, and the size of an integrated combined heat and power plant (CHP).
Each configuration was evaluated based on various objectives: the first objective was the necessary investment costs for the PV system, the battery, and the CHP. The second objective was the yearly total costs from buying electricity from the grid, buying gas to power the CHP, as well as additional maintenance and operating costs.
The third objective was the resilience of the configuration, referring in this case to the amount of time, that a building is able to operate without energy supply from the grid. Note that in the present data set, the resilience is given as a negative value were more negative numbers indicate a higher resilience (lower is better).

\topic{We look at three of the many objectives, dividing them into 2 subspaces}
Within the resulting data set of the pareto-optimal solutions, i.e., solutions that each 
are an optimal trade-off between different constraints, we applied the concept identification algorithm and k-means clustering to identify three different configuration concepts or clusters. We considered three features, the initial \emph{investment costs}, \emph{yearly total costs}, and \emph{resilience}. 
Additionally, we defined two relevant subspaces, a) the \emph{investment costs}, and b) a combination of the expected \emph{yearly total costs} for operation and the solutions' \emph{resilience}.

The total number of parameters that are optimized for the concept identification approach hence amounts to $N_{P} = 3 \cdot (1(1+3)/2 + 2(2+3)/2) = 21$. The CMA-ES is again applied with a population size of 10 and 1000 generations.

\begin{figure*}[ht]
    \centering
    \includegraphics[width=0.96\textwidth]{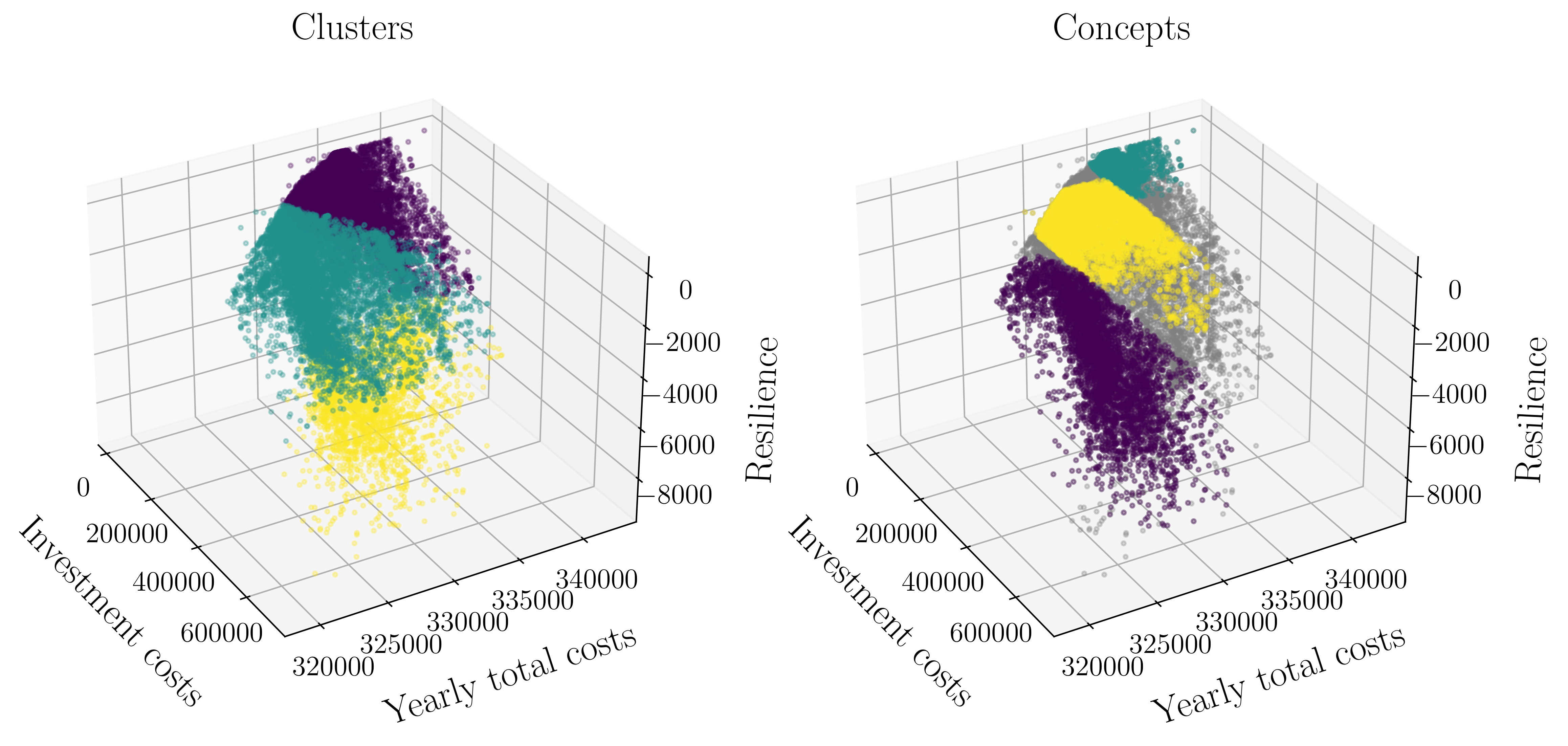}
    \caption{Comparison of clusters found by the k-means algorithm and the identified concepts: The data set of energy management configuration is shown with respect to three objectives. The three clusters and concepts are marked in purple, green, and yellow. In both cases, the groups are distinguishable when all three features are considered as the basis for distinction.}
    \label{fig:Comp_3d}
\end{figure*}

\begin{figure*}[ht]
    \centering
    \includegraphics[width=0.96\textwidth]{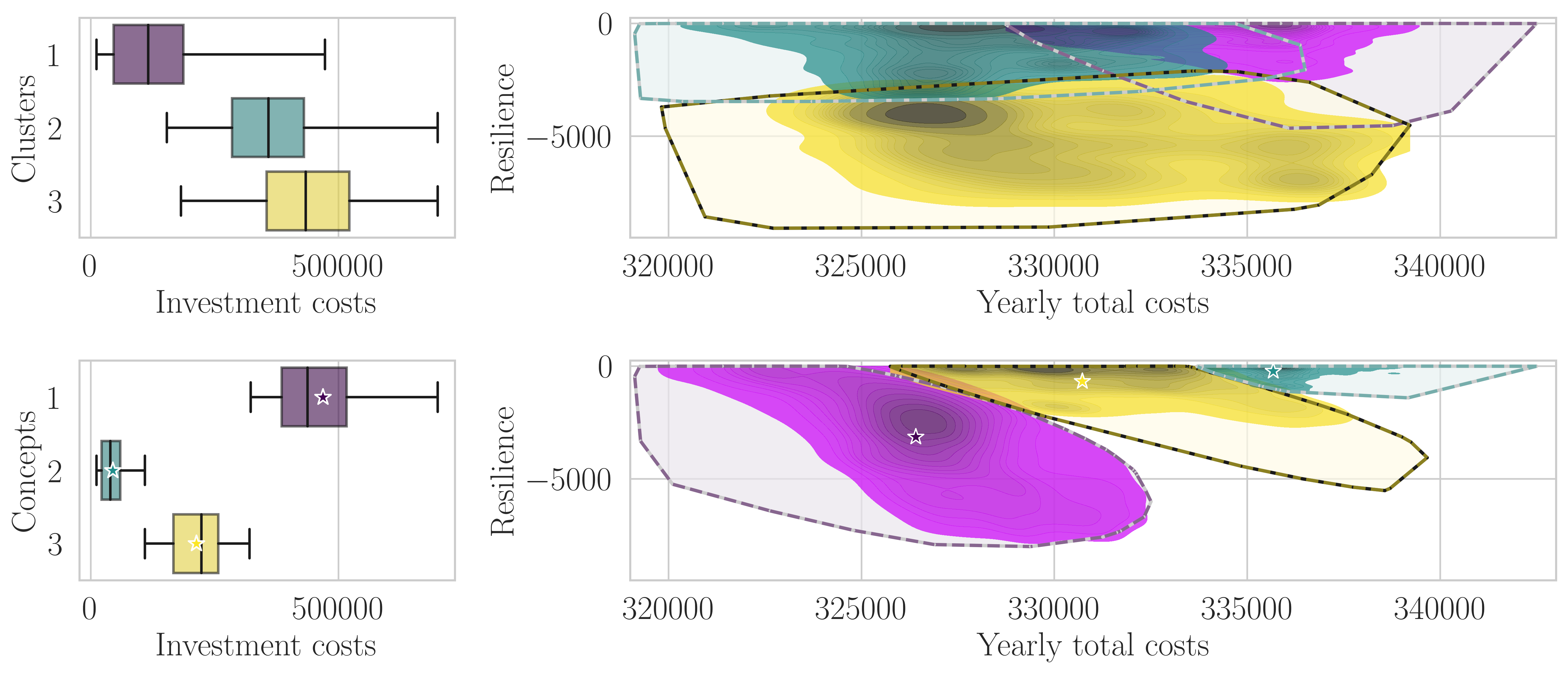}
    \caption{Comparison of clusters found by the k-means algorithm and the identified concepts in the two subspaces of interest: 
    The distribution of the three clusters and concepts (marked purple, green, and yellow) are shown in the one-dimensional space of investment costs. 
    The box indicates the 25-75\%-percentile and the mean, the whiskers show the minimum and maximum values. 
    In the two-dimensional space of resilience over yearly total costs, the groups are shown as the corresponding kernel-density estimation in addition to the convex hull represented by a dashed line. 
    While there is a significant overlap visible between the found clusters, the concept identification leads to a non-overlapping distribution of concepts in both spaces.
    The stars mark selected concept representatives.}
    \label{fig:Comp_3d_2s_kde}
\end{figure*}

\topic{Concept identification works, clustering does not}
As in the previous experiments, both, the clustering algorithm and the concept identification approach lead to distinguishable groups with respect to the full data set (Fig. \ref{fig:Comp_3d}).
However, when projected into the subspaces of interest, a significant overlap of the clusters found by k-means was evident (Fig. \ref{fig:Comp_3d_2s_kde}).
With respect to \emph{investment costs}, all three clusters overlapped, where the green and yellow clusters almost covered the same feature range.
In the second subspace also a significant overlap was visible and a unique allocation of samples to clusters was not possible on the basis of \emph{yearly total costs} and \emph{resilience}.

\topic{The identified concepts are differentiable based on investment and there are apparent trade-off options for yearly total costs and resilience}
On the other hand, the concept identification algorithm identified concepts that were clearly separable within both spaces.
With respect to the first subspace, \emph{investment cost}, the purple, green, and yellow concepts represent high, low, and medium investment solutions, respectively.
Clusters were also unique, i.e., non-overlapping, with respect to the second subspace, presenting trade-off solutions for \emph{yearly total costs} and \emph{resilience}. Within this 
second subspace, inferred clusters comprised solutions that showed an anticorrelated trend, where higher costs were associated with lower resilience values, indicating more robust solutions. 

\topic{Selection of archetypes}
One purpose for the identification of concepts in engineering design tasks is the selection of highly representative instances.
Those instances can be understood as concept archetypes, which, for example, may be used as starting points or prototypes for further refinement of multiple design variations.
From the configuration concepts, we derived representatives by identifying the instance that is closest to the corresponding concept's mean with respect to the full data set (Fig.~\ref{fig:Comp_3d_2s_kde}).

\topic{The decision maker gets nice concepts}
In conclusion, the concept identification process provided the decision maker with different configuration options that meet the initially defined requirements
and thus led to an interpretable clustering solution. 
In particular, clusters were non-overlapping with respect to their investment cost, allowing for a clear separation into three cost levels. Furthermore, 
within each level, solutions followed a trend where higher operation cost led to higher resilience.  
For example, choosing the purple concept allows for realizing configurations with either very low yearly costs, very good resilience values, or a good trade-off between the two (at the cost of a high investment).
Choosing the green concept cannot achieve the same performance with respect to yearly costs and resilience, is however realizable with a smaller investment budget.
The yellow concept is essentially a trade-off concept in between the other two.
We conclude that enforcing non-overlapping solutions that are also clearly separable within relevant feature subspaces may aid decision-making processes as solutions 
allow for a clearer interpretation in terms of relevant feature types.

\section{Discussion}\label{sec:discussion}
\topic{Finding design concepts is a relevant task in engineering design processes}
The identification of designs sharing similar characteristics in different meaningful description spaces is a central component of the engineering design process.
In the present paper, we propose that this concept identification problem is a special form of clustering problem, where additional constraints are 
imposed on the clustering solution. These constraints are---besides identifying dense and well-separated clusters---a non-overlap of clusters and 
that non-overlapping clusters persist when only subsets of features are considered.
We here demonstrate that a recently proposed concept identification algorithm returns such clustering solutions and may be applicable to problems 
outside of the engineering design domain.

\topic{The experiments showed three things}
In three experiments, we show that this recently proposed concept identification algorithm fulfills the three constraints, which is not achieved by 
classical clustering algorithms. We further introduce the MI as an additional evaluation metric, next to classical evaluation metrics for 
clustering solutions, which specifically tests whether identified solutions comprise consistent clusters across subspaces. Last, we demonstrate the 
application of the concept identification algorithm in a decision making problem on optimization results from the energy management domain, where  
the concept identification algorithm identified more interpretable clusters and thus eased the decision-making process.

\topic{Concept identification is a new clustering algorithm that meets specific constraints and generates non-overlapping and consistent clusters}
From the fact that the distribution of identified concepts demonstrates higher MI values for the two artificial data sets and the energy management application scenario, we conclude that the proposed method is well suited to find concepts with high consistency across defined subspaces.
The method can be set up to identify groups that share similar properties based on an arbitrary combination of subspaces (i.e. subsets of features).
The experiments further show, that traditional clustering methods do not lead to solutions with similar properties.
The proposed clusters showed significant overlap in the subspaces, hence providing no consistent solution across subspaces.
Consistency may be a desirable property in additional applications, such as cross-domain recommender systems.

\topic{Mutual Information can be used to evaluate the quality of clusters and concepts based on their consistency across multiple subspaces.}
Aside from a visual inspection, the quality of concepts can be evaluated by estimating the MI between features defining the subspaces.
The MI provides an assessment of consistency, essentially evaluating how similar the features of sample groups are in predefined subspaces of the full feature set.
In all experiments, the MI was persistently higher for the solution returned by the concept identification algorithm, compared to the clusters found by traditional clustering algorithms.

\topic{Mutual Information cannot directly be used as an optimization criterion for concept identification of clustering (in this case), because it cannot be used to generate non-overlapping clusters}
MI proved to be a good estimate for concept quality by evaluating consistency across multiple subspaces. 
This, however, leads to the question if MI can directly be used to optimize the distribution of concepts.
Instead of evaluating the performance of a separate concept identification method, one option would be to implement an estimation of the MI as the evaluation function of a concept optimization algorithm.
Such a process leads to groups of data samples that are optimized based on their consistency across multiple subspaces.
However, the found groups are likely to be indistinguishable as concepts---while a process that optimizes MI across multiple feature sets can assure consistency, it neglects the two other requirements for reasonable clusters: compactness and uniqueness (i.e. non-overlapping groups).
An assignment of samples into highly overlapping groups is not suitable for the identification of concepts, as the groups are likely to be too similar to function as distinct alternatives.
The benefit for the user would hence be very limited.

\topic{How to choose sensible subspaces remains a difficult tasks for the engineer}
While the identification process provides the user with consistent clusters, the choice of subspaces in the first place remains a challenging task.
A sensible split of features into separate subspaces requires in-depth domain knowledge, aside from a thorough understanding of the application domain.
The choice of subspaces has a high influence on the potential identification outcome and defines the constitution of a concept in the particular task.

\topic{Choosing useful representatives is also difficult}
Another topic that allows for further research is the choice of concept representatives.
In many application scenarios, the user wants to select a manageable amount of individual samples from each concept (which can consist of several thousands of samples or more).
Depending on the design task, various choices are reasonable.
If, for example, the concept identification is part of an iterative optimization process, it is sensible to pick archetypal configurations from each concept as prototypes for further optimization steps. 
Samples that are closest to the mean of the full feature vector can be understood as archetypes for each concept.
But a selection of representatives based on fewer features, or an entire different criterion can be equally reasonable and depends on the engineering task.

\section{Conclusion}\label{sec:conclusion}

Concept identification, originally proposed in the engineering design domain, may be understood as a special form of clustering problem. 
Thus, concept identification algorithms may be applied as clustering algorithms with properties that are relevant to application domains other than engineering design. 
In particular, solutions comprise a consistent clustering of instances across the full feature space and a-priori defined subspaces that are relevant to the specific problem. 
We demonstrate how concept identification solutions differ from classical clustering solutions on two artificial data sets, designed to illustrate these differences. 
Further, we introduce the Mutual Information (MI) as an evaluation criterion of the concept identification algorithm. 
The MI quantifies this consistency across subspaces for a given clustering solution.
We conclude that the MI is a suitable metric to compare various clustering approaches, when consistency across subspaces is required.
Last, we show how concept identification algorithms may be used in application 
domains other than engineering design, and successfully apply the algorithm in a decision-making task from energy 
management optimization. 

\topic{With the experiments we demonstrate that concept identification can meet the constraints of the task 
and that clustering does not}
In three experiments we demonstrated that a) concept identification leads to consistent clusters in selectable 
subspaces and b) common clustering algorithms do not have this property.
From this, we further conclude that concept identification can generally be understood as a clustering technique 
that not only generates compact and clearly distinguishable groups, but also allows to integrate a third clustering 
objective, namely, ensuring the consistency of the groups with respect to predefined subspaces of the data.

\topic{Final conclusion}
In conclusion, we propose concept identification algorithms as a suitable tool to generate clustering solutions with desirable properties, such as increased interpretability and explainability. 
We further introduce Mutual Information as a suitable metric to assess these properties.


\bibliographystyle{IEEEtran}
\bibliography{bib_mi}

\begin{thebibliography}{10}
\providecommand{\url}[1]{#1}
\csname url@samestyle\endcsname
\providecommand{\newblock}{\relax}
\providecommand{\bibinfo}[2]{#2}
\providecommand{\BIBentrySTDinterwordspacing}{\spaceskip=0pt\relax}
\providecommand{\BIBentryALTinterwordstretchfactor}{4}
\providecommand{\BIBentryALTinterwordspacing}{\spaceskip=\fontdimen2\font plus
\BIBentryALTinterwordstretchfactor\fontdimen3\font minus
  \fontdimen4\font\relax}
\providecommand{\BIBforeignlanguage}[2]{{%
\expandafter\ifx\csname l@#1\endcsname\relax
\typeout{** WARNING: IEEEtran.bst: No hyphenation pattern has been}%
\typeout{** loaded for the language `#1'. Using the pattern for}%
\typeout{** the default language instead.}%
\else
\language=\csname l@#1\endcsname
\fi
#2}}
\providecommand{\BIBdecl}{\relax}
\BIBdecl

\bibitem{Lanfermann2020}
F.~Lanfermann, S.~Schmitt, and S.~Menzel, ``{An Effective Measure to Identify
  Meaningful Concepts in Engineering Design optimization},'' in \emph{2020 IEEE
  Symposium Series on Computational Intelligence (SSCI)}.\hskip 1em plus 0.5em
  minus 0.4em\relax IEEE, dec 2020, pp. 934--941.

\bibitem{Lanfermann2022}
F.~Lanfermann and S.~Schmitt, ``Concept identification for complex engineering
  datasets,'' \emph{Advanced Engineering Informatics}, vol.~53, p. 101704,
  2022.

\bibitem{Khan2017}
M.~M. Khan, R.~Ibrahim, and I.~Ghani, ``Cross domain recommender systems: a
  systematic literature review,'' \emph{ACM Computing Surveys (CSUR)}, vol.~50,
  no.~3, pp. 1--34, 2017.

\bibitem{Hastie2009}
T.~Hastie, R.~Tibshirani, and J.~Friedman, \emph{The Elements of Statistical
  Learning}, 2nd~ed.\hskip 1em plus 0.5em minus 0.4em\relax New York: Springer,
  2009.

\bibitem{Liu2010}
Y.~Liu, Z.~Li, H.~Xiong, X.~Gao, and J.~Wu, ``Understanding of internal
  clustering validation measures,'' in \emph{2010 IEEE International Conference
  on Data Mining}.\hskip 1em plus 0.5em minus 0.4em\relax IEEE, 2010, pp.
  911--916.

\bibitem{Parsons2004}
L.~Parsons, E.~Haque, and H.~Liu, ``{Subspace clustering for high dimensional
  data},'' \emph{ACM SIGKDD Explorations Newsletter}, vol.~6, no.~1, pp.
  90--105, jun 2004.

\bibitem{Bickel2004}
S.~Bickel and T.~Scheffer, ``{Multi-View Clustering},'' in \emph{Fourth IEEE
  International Conference on Data Mining (ICDM'04)}.\hskip 1em plus 0.5em
  minus 0.4em\relax IEEE, 2004, pp. 19--26.

\bibitem{Tan2000}
P.~Tan and V.~Kumar, ``{Interestingness Measures for Association Patterns: A
  Perspective},'' in \emph{KDD Workshop on Postprocessing in Machine Learning
  and Data Mining}, 2000.

\bibitem{Graening2014}
L.~Graening and B.~Sendhoff, ``{Shape Mining: A Holistic Data Mining Approach
  for Engineering Design},'' \emph{Advanced Engineering Informatics}, vol.~28,
  no.~2, pp. 166--185, 2014.

\bibitem{MacQueen1967}
J.~MacQueen, ``{Some methods for classification and analysis of multivariate
  observations},'' \emph{Proceedings of the Fifth Berkeley Symposium on
  Mathematical Statistics and Probability}, vol.~1, pp. 281----297, 1967.

\bibitem{Arthur2007}
D.~Arthur and S.~Vassilvitskii, ``{k-means++: The Advantages of Careful
  Seeding},'' \emph{Proceedings of the eighteenth annual ACM-SIAM symposium on
  Discrete algorithms}, pp. 1027--1035, 2007.

\bibitem{Bezdek1981}
J.~C. Bezdek, \emph{{Pattern Recognition with Fuzzy Objective Function
  Algorithms}}.\hskip 1em plus 0.5em minus 0.4em\relax Boston, MA: Springer US,
  1981.

\bibitem{Rasmussen2000}
C.~E. Rasmussen, ``{The infinite Gaussian mixture model},'' \emph{Advances in
  Neural Information Processing Systems}, no.~1, pp. 554--559, 2000.

\bibitem{Sim2013}
K.~Sim, V.~Gopalkrishnan, A.~Zimek, and G.~Cong, ``{A survey on enhanced
  subspace clustering},'' \emph{Data Mining and Knowledge Discovery}, vol.~26,
  no.~2, pp. 332--397, 2013.

\bibitem{Ntelemis2022}
F.~Ntelemis, Y.~Jin, and S.~A. Thomas, ``{Information maximization clustering
  via multi-view self-labelling},'' \emph{Knowledge-Based Systems}, p. 109042,
  2022.

\bibitem{Yang2018}
Y.~Yang and H.~Wang, ``{Multi-view clustering: A survey},'' \emph{Big Data
  Mining and Analytics}, vol.~1, no.~2, pp. 83--107, jun 2018.

\bibitem{Cai2013}
X.~Cai, F.~Nie, and H.~Huang, ``{Multi-View K-Means Clustering on Big Data
  Xiao},'' in \emph{Proceedings of the Twenty-Third International Joint
  Conference on Artificial Intelligence}, sep 2013.

\bibitem{Gao2015}
H.~Gao, F.~Nie, X.~Li, and H.~Huang, ``{Multi-view Subspace Clustering},'' in
  \emph{2015 IEEE International Conference on Computer Vision (ICCV)}.\hskip
  1em plus 0.5em minus 0.4em\relax IEEE, dec 2015, pp. 4238--4246.

\bibitem{Rousseeuw1987}
P.~J. Rousseeuw, ``{Silhouettes: A graphical aid to the interpretation and
  validation of cluster analysis},'' \emph{Journal of Computational and Applied
  Mathematics}, vol.~20, no.~C, pp. 53--65, 1987.

\bibitem{Shannon1948}
C.~E. Shannon, ``{A mathematical theory of communication},'' \emph{The Bell
  System Technical Journal}, vol.~27, pp. 379--423, 1948.

\bibitem{Davies1979}
D.~L. Davies and D.~W. Bouldin, ``A cluster separation measure,'' \emph{IEEE
  Transactions on Pattern Analysis and Machine Intelligence}, vol. PAMI-1,
  no.~2, pp. 224--227, 1979.

\bibitem{MacKay2005}
D.~J.~C. MacKay, \emph{{Information Theory, Inference, and Learning
  Algorithms}}.\hskip 1em plus 0.5em minus 0.4em\relax Cambridge, UK: Cambridge
  University Press, 2005.

\bibitem{Kraskov2004}
A.~Kraskov, H.~St{\"{o}}gbauer, and P.~Grassberger, ``{Estimating mutual
  information},'' \emph{Physical Review E}, vol.~69, no.~6, p.~16, 2004.

\bibitem{Wollstadt2019}
\BIBentryALTinterwordspacing
P.~Wollstadt, J.~T. Lizier, R.~Vicente, C.~Finn, M.~Mart\'inez-Zarzuela,
  P.~A.~M. Mediano, L.~Novelli, and M.~Wibral, ``{IDTxl: The Information
  Dynamics Toolkit xl: a Python package for the efficient analysis of
  multivariate information dynamics in networks},'' \emph{Journal of Open
  Source Software}, vol.~4, no.~34, p. 1081, 2019. [Online]. Available:
  \url{https://github.com/pwollstadt/IDTxl}
\BIBentrySTDinterwordspacing

\bibitem{Lizier2014}
J.~T. Lizier, ``{{JIDT}}: An information-theoretic toolkit for studying the
  dynamics of complex systems,'' \emph{Frontiers in Robotics and AI}, vol.~1,
  p.~11, 2014.

\bibitem{Paninski2003}
L.~Paninski, ``Estimation of entropy and mutual information,'' \emph{Neural
  Computation}, vol.~15, no.~6, pp. 1191--1253, 2003.

\bibitem{Vicente2011}
R.~Vicente, M.~Wibral, M.~Lindner, and G.~Pipa, ``{Transfer entropy-a
  model-free measure of effective connectivity for the neurosciences},''
  \emph{Journal of Computational Neuroscience}, vol.~30, no.~1, pp. 45--67,
  2011.

\bibitem{hansenCMA}
N.~Hansen, \emph{The CMA Evolution Strategy: A Comparing Review}.\hskip 1em
  plus 0.5em minus 0.4em\relax Berlin, Heidelberg: Springer Berlin Heidelberg,
  2006, pp. 75--102.

\bibitem{Hansen2001}
N.~Hansen and A.~Ostermeier, ``{Completely Derandomized Self-Adaptation in
  Evolution Strategies},'' \emph{Evolutionary Computation}, vol.~9, no.~2, pp.
  159--195, 2001.

\bibitem{Liu2022}
Q.~Liu, F.~Lanfermann, T.~Rodemann, M.~Olhofer, and Y.~Jin,
  ``Surrogate-assisted many-objective optimization of building energy
  management,'' \emph{IEEE Computational Intelligence Magazine}, vol.~18,
  no.~4, pp. 14--28, 2023.

\end{thebibliography}

\end{document}